\documentclass[12pt]{article}
\usepackage[utf8]{inputenc}
\usepackage{graphicx}
\usepackage[a4paper, total={7in, 8in}]{geometry}
\usepackage{booktabs}
\usepackage{titling}
\usepackage{indentfirst}
\usepackage{titlesec}
\usepackage{booktabs}
\usepackage{natbib}
\usepackage{setspace}
\usepackage[hyperfootnotes=false]{hyperref}
\usepackage{tablefootnote}
\usepackage{float}
\usepackage{array}
\usepackage{titling}
\usepackage{tikz}
\usetikzlibrary{shapes.geometric, arrows}
\usepackage{adjustbox}
\usepackage{hyperref}

\tikzstyle{process} = [rectangle, 
minimum width=7.2cm, 
minimum height=1cm, 
align=left,
text width=6.9cm, 
draw=black, 
fill=orange!10]

\tikzstyle{arrow} = [thick,->,>=stealth]

\title{Automated Annotation with Generative AI Requires Validation}
\author{Nicholas Pangakis\thanks{Department of Political Science, University of Pennsylvania}, Samuel Wolken\thanks{Department of Political Science and Annenberg School for Communication, University of Pennsylvania}, and Neil Fasching\thanks{Annenberg School for Communication, University of Pennsylvania}}

\begin{document}

\maketitle
\thispagestyle{empty}
\renewcommand{\thefootnote}{}

\begin{abstract}
	Generative large language models (LLMs) can be a powerful tool for augmenting text annotation procedures, but their performance varies across annotation tasks due to prompt quality, text data idiosyncrasies, and conceptual difficulty. Because these challenges will persist even as LLM technology improves, we argue that \textit{any} automated annotation process using an LLM must validate the LLM’s performance against labels generated by humans. To this end, we outline a workflow to harness the annotation potential of LLMs in a principled, efficient way. Using GPT-4, we validate this approach by replicating 27 annotation tasks across 11 datasets from recent social science articles in high-impact journals. We find that LLM performance for text annotation is promising but highly contingent on both the dataset and the type of annotation task, which reinforces the necessity to validate on a task-by-task basis. We make available easy-to-use software designed to implement our workflow and streamline the deployment of LLMs for automated annotation.\footnote{Author contributions: NP and SW developed the code, analyzed the data, and wrote the manuscript. NP, SW, and NF collected data and edited the manuscript.}\footnote{We thank Yphtach Lelkes for his support and guidance and Daniel Hopkins for his helpful comments. We also thank any author that sent us their data.}

\end{abstract}

\pagebreak


\section{Introduction}

\renewcommand{\thefootnote}{\arabic{footnote}}
\setcounter{footnote}{0}

Many tasks in natural language processing (NLP) depend on high-quality, manually-labeled text data for training and validation. The manual annotation process, however, poses non-trivial challenges. In addition to being time consuming and expensive, human annotators, usually crowdsourced workers\footnote{Researchers have raised concerns about the quality of data collected from workers on popular crowdsourcing platforms like Amazon's Mechanical Turk \citep{Chmielewski20, Douglas23}} or undergraduate assistants, often suffer from a limited attention span, fatigue, and changing perceptions of the underlying conceptual categories throughout the labeling procedures \citep{grimmer2013, Neuendorf16}. When labeling large amounts of data, these limitations can lead to labeled text data that suffer from inconsistencies and errors that may be unobservable and correlated—especially when using coders with similar demographic backgrounds.

To address these challenges, researchers have recently explored the potential of generative large language models (LLMs), such as ChatGPT, to replace human annotators. LLMs are faster, cheaper, reproducible, and not susceptible to some of the pitfalls in human annotation. Prior applications using LLMs in this capacity, however, have yielded mixed results. \citet{gilardi2023chatgpt} claim that ChatGPT outperforms MTurkers on a variety of annotation tasks. \citet{reiss23}, on the other hand, finds that LLMs perform poorly on text annotation and argues that this tool should be used cautiously. Numerous other studies present similar analyses with varying results and recommendations \citep[e.g.,][]{he2023annollm, wang-etal-2021-want-reduce, ziems23, zhu2023chatgpt, ding2022gpt3}.

Although existing research provides valuable insights into both the benefits and potential constraints of utilizing LLMs for text annotation, it is unclear whether the automated labeling workflows used in these studies can be confidently applied to other datasets and tasks, given that they report minimal performance metrics (i.e., only accuracy or intercoder agreement) and analyze a few insular tasks on a small number of datasets. Moreover, due to the rapid development of LLM technology, it is likely that any binary recommendations about the competence of LLMs at one-off tasks will be quickly outdated with additional LLM advancements. While several studies so far have applied LLMs to a broader range of datasets \citep{ziems23, zhu2023chatgpt}, they exclusively test LLM annotation performance on popular, publicly available benchmark datasets. As such, these tests are plausibly affected by contamination \citep{NEURIPS2020_1457c0d6}, meaning that the datasets may be included in the LLM's training data and that strong performance may reflect memorization, which will not generalize to new datasets and tasks. Without clear guidance on a recommended workflow to use these tools safely and effectively, academics and practitioners alike could deploy this type of tool when it has suboptimal performance. 

Our primary argument is that a researcher using an LLM for automated annotation must \textit{always} validate the LLM's performance against a subset of high-quality human-annotated labels.\footnote{By high quality, we refer to labels generated by subject matter experts (not undergraduate assistants or crowdsourced workers) who carefully label a small subset of data. As we describe below, our recommended workflow only requires a fraction of the labeled text data when compared to existing procedures.} While LLMs can be an effective tool to augment the annotation process, there are circumstances where LLMs fail to deliver accurate results due to factors such as ineffective prompts, noisy text data, and difficult annotation tasks. Because these challenges will persist even as LLM technology improves,\footnote{For example, even as LLM technology increases in sophistication, humans could still provide ambiguous prompt instructions, which could negatively affect LLM performance on an annotation task.} researchers augmenting their text annotation approach with LLMs must always validate on a task-by-task basis. Rigorous validation can help researchers craft effective prompts and determine whether LLM annotation is viable for their classification tasks and datasets. To this end, we outline a recommended workflow to harness the potential of LLMs for text annotation in a principled, efficient way. 

In this paper, we propose and test a workflow for augmenting and automating annotation projects with LLMs. We validate our approach and test the capabilities of LLMs on a wide range of annotation tasks from different, non-public datasets using appropriate performance metrics. We use LLMs to replicate 27 different annotation processes from 11 non-public datasets used from articles recently featured in high-impact publications. In total, we classified over 200,000 text samples using an LLM (i.e., GPT-4).\footnote{Scholars have raised concerns about relying on LLMs in social science research due to the constant evolution of LLM software, the black-box nature of how LLMs process queries, and ambiguity about the training data that underlies LLMs. Open-source LLMs represent a viable solution to many of these problems \citep{spirling2023}. The workflow outlined here is LLM-agnostic and could be adapted to any open-source LLM.}

Our findings indicate that LLM performance for text annotation is promising but highly contingent on both the dataset and the type of annotation task. Across all tasks, we report a median accuracy of 0.850 and a median F1 of 0.707. Despite the strong overall performance, nine of the 27 tasks had either precision or recall
below 0.5, which reinforces the necessity of researcher validation. Given the variation in performance across tasks, we identify four different use cases for automated annotation procedures guided by the LLMs validation performance. These include using an LLM to check the quality of human-labeled data, using an LLM to identify cases to prioritize for human review, using an LLM to produce labeled data to finetune and validate a supervised classifier, and using an LLM to classify an entire text corpus.

In addition to recommending a standardized process to validate \textit{when} and \textit{how} to use LLMs, we also introduce several novel tools in our automated annotation procedures.\footnote{Code available here: \url{https://github.com/npangakis/gpt_annotate}} First, we make available easy-to-use software in Python designed to implement our methods and streamline the deployment of LLMs for automated annotation. Second, we show the utility of a \textit{consistency score}. To measure how consistently an LLM predicts a particular text sample’s label, we repeatedly classify each text sample at an LLM temperature of 0.6. Treating the modal answer as the predicted LLM label, we approximate a degree of “consistency” across each LLM classification. Because there is a strong correlation between higher consistency scores and the probability of the classification being correct, consistency scores are an effective way for researchers to identify edge cases.

\section{Workflow and validation}

\begin{figure}
    \centering
    \begin{tikzpicture}[node distance=3cm]

\node (step1) [process] {\textbf{Step 1}: Researcher creates task-specific instructions (i.e., a codebook).};

\node (step2) [process, below of=step1] {\textbf{Step 2}: Using codebook, subject matter experts annotate random subset of text samples.};

\node (step3) [process, below of =step2] {\textbf{Step 3}: Use LLM to annotate a subset of the human-labeled data
using the same codebook. Then, evaluate performance
by comparing the LLM labels against the
human labels.};

\node (step4) [process, right of=step3, xshift=6cm] {\textbf{Step 4}: If low performance, refine codebook to
emphasize incorrect classifications. If necessary, repeat steps 2 and 3 with updated codebook.};

\node (step5) [process, below of=step3] {\textbf{Step 5}: Using final codebook, test LLM performance on remaining human-labeled samples.};

\draw [arrow] (step1) -- (step2);
\draw [arrow] (step2) -- (step3);
\draw [arrow] (step3) -- (step4);
\draw [arrow] (step4) -- (step3);
\draw [arrow] (step3) -- (step5);

\end{tikzpicture}
    \caption{Workflow for augmenting text annotation with an LLM}
    \label{fig:workflow}
\end{figure}
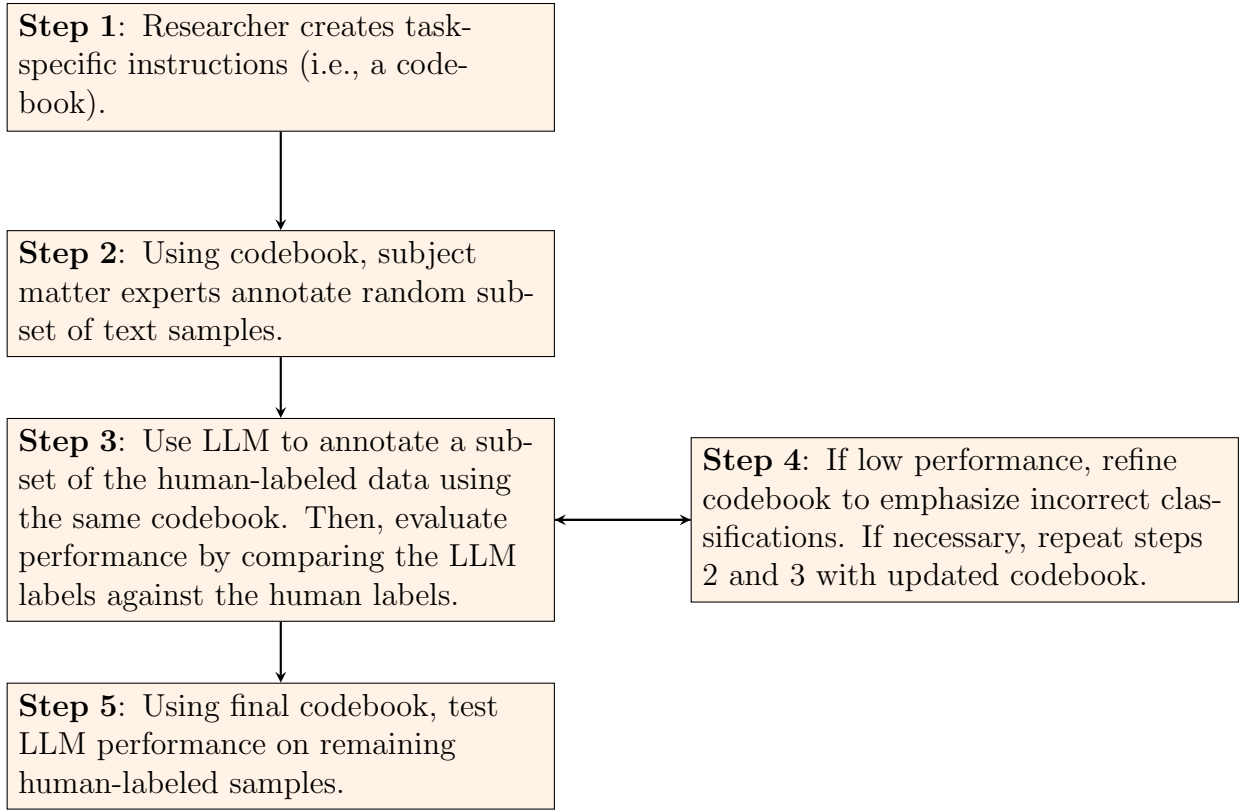

In the absence of clear guidance on a recommended workflow for utilizing LLMs for annotation tasks, researchers run the risk of deploying these tools despite poor performance. Displayed in Figure \ref{fig:workflow}, we offer a five-step workflow for incorporating LLMs in a way that foregrounds human judgment, includes an opportunity for human-in-the-loop refinement of instructions, and gives a clear indication of LLM performance with minimal ex-ante investment of resources. We designed this workflow with two motivations: to validate LLM performance and to refine prompts for LLM classification when possible. 

As Figure \ref{fig:workflow} shows, researchers should first craft a set of task-specific instructions (i.e., a codebook)\footnote{Because LLMs respond to natural language prompts, a clear codebook is essential for LLM annotation tasks. In general, an annotation codebook should clearly delineate the concepts of interest. A rich literature spanning both qualitative and quantitative social science offers guidance on how to develop a codebook and classify text data based on relevant concepts \citep[e.g., ][]{krip, crabtree92, doi:10.1177/1525822X980100020301}.} and then have at least two subject matter experts and an LLM annotate the same text samples—with the sample size depending on the type of task and class imbalance.\footnote{Based on our analyses, the number of samples annotated should range between 250 and 1,250 random text samples. For annotation tasks that involve identifying very rare cases (for instance, cases in which fewer than 1 percent of text samples are coded as positive cases for a particular dimension), it may be necessary to label significantly more text samples by hand.} Critically, both the human coders and the LLM should use the same codebook for annotation, where the codebook serves as the LLM's prompt instructions.\footnote{If humans and LLMs label the data using different codebooks, there could be a conceptual gap between the two annotation instructions.} Then, the researcher should evaluate performance (i.e., accuracy, recall, precision, and F1) by comparing the predicted LLM labels against the human labels. As an initial step, researchers should first have the LLM annotate a subset of the human-labeled text samples. If LLM performance is low on this subset, researchers can refine the codebook instructions by emphasizing incorrect classifications (repeating this process if necessary).\footnote{After updating the codebook, researchers may want to use the updated codebook to classify the same subset of cases again to measure the magnitude of change their edits caused. This exercise is overfitting to a particular subset of data and thus should not be taken as the barometer of performance. Instead, it offers a straightforward gauge of how significantly the codebook edits affect LLM decision-making. If fundamental changes were made to the codebook, the researcher may want to have the humans re-label the text samples as well.} Finally, using the updated codebook, researchers should test the LLM performance against the remaining human-labeled samples. Performance on this held-out data should be used to determine whether an LLM can be effectively utilized for automated annotation. The code made available in our GitHub repository offers a simple and efficient way to implement the procedures outlined in the above workflow.

To validate the proposed workflow, we use an LLM to replicate 27 annotation tasks across 11 datasets. To ensure these tasks represent a range of annotation tasks in contemporary social science research, we draw from research published in outlets across a spectrum of disciplines ranging from interdisciplinary publications (e.g., \textit{Science Advances} and \textit{Proceedings of the National Academy of Sciences}) to high-impact field journals in political science (e.g., \textit{American Political Science Review} and \textit{American Journal of Political Science}) and psychology (e.g., \textit{Journal of Personality and Social Psychology}). In the Appendix, Table \ref{tab:articles} includes the complete list of replication articles and Table \ref{tab:articles2} provides brief descriptions of the annotation tasks in this articles.\footnote{To find these articles, we searched high-impact journals for articles that implemented some type of manual annotation procedure. If we were able to acquire the text data, we replicated all annotation procedures from articles that were published within the last three years.} These annotation tasks cover an extensive range of social science applications, from identifying whether Cold War-era texts pertained to foreign affairs or military matters \citep{schub22} to analyzing open-ended survey responses to classify how people conceptualize where beliefs come from \citep{Cusimano20}.

In each case, we replicate an annotation task using the human-labeled data from the original study as the ground truth. To avoid the potential for contamination, we rely exclusively on datasets stored in password-protected data archives (e.g., Dataverse) or datasets secured through direct outreach to authors.\footnote{To harmonize this diverse range of annotation tasks into a common framework for evaluation, we treat every dimension as a separate binary annotation task. Thus, if an article includes a classification task with three potential labels, we split the annotation process into three discrete binary classification tasks.} Whenever possible, we begin with the exact codebook used in the original research design. If this codebook is not available, we either quote or paraphrase text from the article or supplementary materials that describes the concepts of interest.\footnote{We do not observe any relationship between LLM performance and whether or not the direct codebook was available.}

Across all 27 tasks, we annotate slightly over 200,000 text samples using OpenAI's \textsc{GPT-4} API. The overall cost was approximately \$420 USD. On average, a dataset with 1,000 text samples took approximately 2--3 hours to complete seven iterations (see ``Consistency scores'' section below). Together, the low cost and relatively rapid speed demonstrate the potential value of LLM-augmented annotation for many social science text analysis tasks.   

\section{Results}

\begin{figure}
    \centering
    \includegraphics{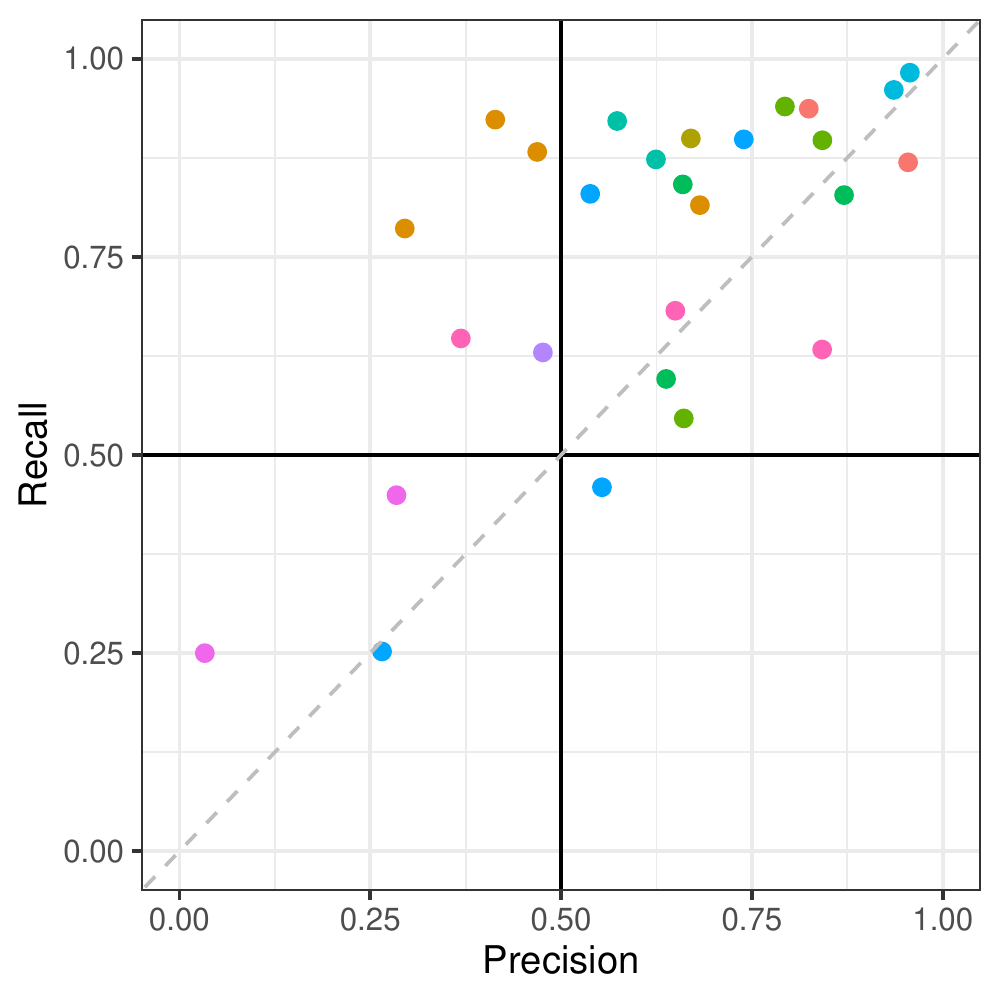}
    \caption{Precision and recall for each of 27 replicated classification tasks. Color reflects dataset, such that points sharing the same color are conducted on the same text data.}
    \label{fig:quadrant}
\end{figure}

Classification results are shown in Table \ref{tab:metrics}. The results reported here are based on ``held out'' text samples (i.e., not the text samples used in the codebook update process from Step 4 of the workflow). Across the 27 tasks, LLM classification performance achieved a median F1 score of 0.707. Figure \ref{fig:quadrant} shows performance on precision and recall for each classification task. As is apparent in this figure, LLM classification performance is stronger in recall than precision for 20 of the 27 tasks. On eight of the 27 tasks, the LLM achieves remarkably strong performance with precision and recall both exceeding 0.7.

\begin{table}[H]
    \centering
    \begin{tabular}{lccccccc}
    \toprule
    Metric & Minimum & 25th percentile & Mean & Median & 75th percentile   & Maximum \\
    \midrule
    Accuracy &  0.674 & 0.808 & 0.855 &  0.85 & 0.905 & 0.981 \\
    Precision & 0.033 & 0.472 & 0.615 &  0.650 & 0.809 & 0.957 \\
    Recall & 0.25 & 0.631 & 0.749 & 0.829 & 0.899 & 0.982 \\
    F1 & 0.059 &  0.557 & 0.660 &  0.707 & 0.830 & 0.969 \\
    \bottomrule
    \end{tabular}
    \caption{LLM classification performance across 27 tasks from 11 datasets.}
    \label{tab:metrics}
\end{table}

Despite the strong overall performance, nine of the 27 tasks had either precision or recall below 0.5---and three tasks had both precision and recall below 0.5. Thus, for a full one-third of tasks, the LLM either missed at least half of the true positive cases, had more false positives than true positives, or both. As shown in Table \ref{fig:quadrant}, the aggregate performance ranged as low as an F1 score of 0.06. Moreover, LLM performance varied substantially across tasks within a single dataset. In the most extreme example, F1 ranged from 0.259 to 0.811 on two separate tasks within \cite{card23}, a difference of 0.552. These results demonstrate the variability of LLM annotation and, accordingly, underscore the need for task-specific validation. 

\subsection{Consistency scores}

\begin{figure}[H]
    \centering
    \includegraphics[width=0.8\linewidth]{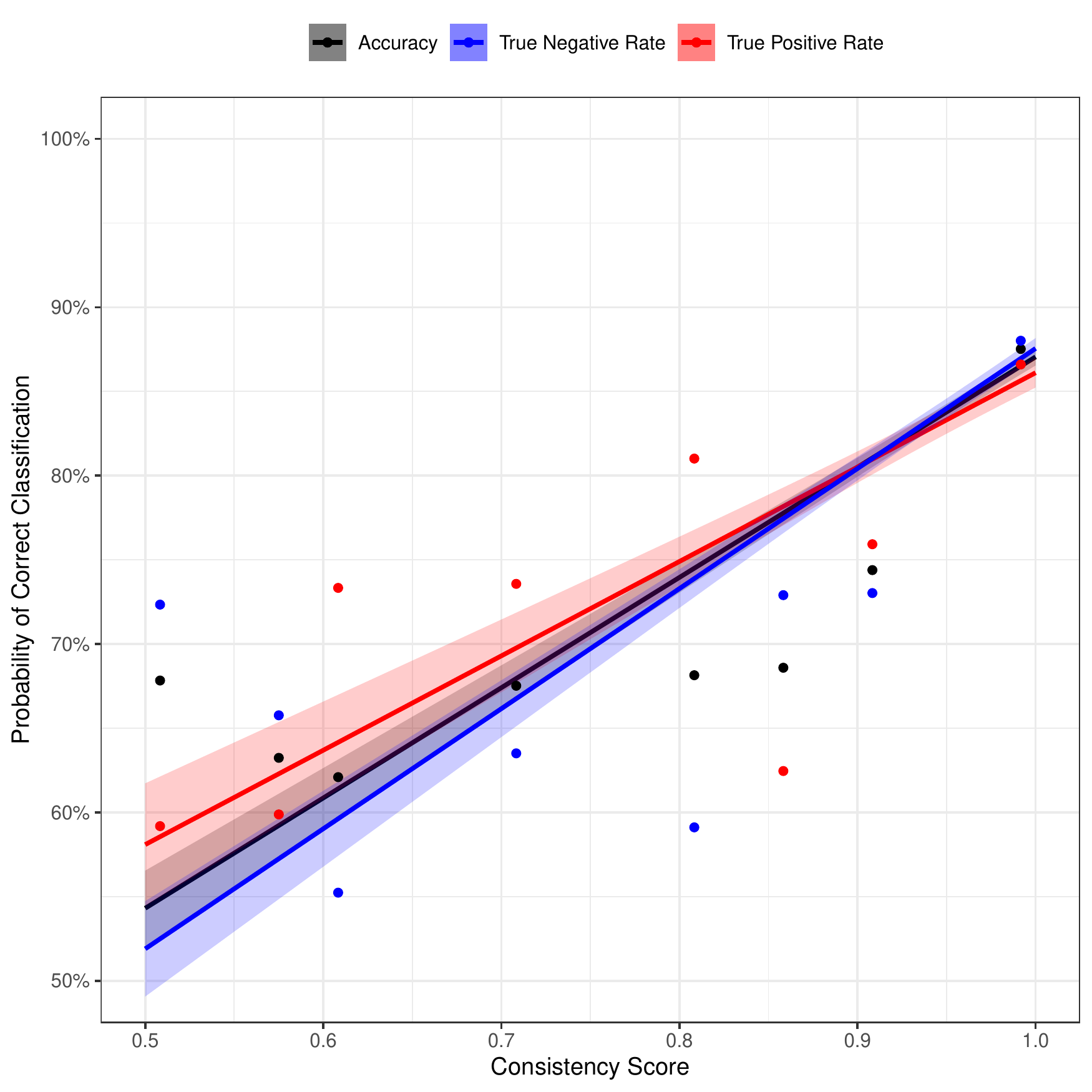}
    \caption{Relationship between consistency score and accuracy, TPR, and TNR.}
    \label{fig:consistency}
\end{figure}

By inducing randomness in the LLM through the use of the temperature hyperparameter and by repeating the annotation task, we can generate an empirical measure of variance in the label that we deem a ``consistency score.'' We recommend that the researcher have the LLM classify each sample at least three times with a temperature above 0.\footnote{Temperature is an LLM hyperparameter that indicates the degree of diversity introduced across LLM responses. It ranges between 0 and 1.} For our analyses, we use a temperature of 0.6 to label each text sample a minimum of seven times with the same codebook.\footnote{While our choice of 0.6 is arbitrary, we did confirm that annotating numerous text samples repeatedly returns better results than a single classification at temperature 0. Future work could test which temperature setting returns optimal results.} As shown in Figure \ref{fig:consistency}, consistency score is correlated with accuracy, true positive rate, and true negative rate. Given a vector of classifications, $C$, with length $l$ for a given classification task, consistency is measured as the proportion of classifications that match the modal classification $\big( \frac{1}{l}\sum_{i=1}^l C_i == C_{mode} \big)$. Across tasks, accuracy is 19.4 percentage points higher for text samples labeled with a consistency score of 1.0 compared with those labeled with a consistency score less than 1.0. The true positive rate and true negative rate are 16.4 percentage points and 21.4 percentage points higher, respectively, for fully consistent classifications. Of all the labeled samples, 85.1\% have a consistency score of 1.0. Therefore, consistency scores offer a useful way of identifying edge cases or more difficult annotations.

\subsection{Codebook updates}

For each of the 27 classification tasks, we follow the previously outlined workflow. In steps 3 and 4 of our workflow, we tested LLM classification performance on a subset of data and then, if relevant, made iterative human-in-the-loop updates to the codebook to optimize the prompt for LLM classification performance. This step can be thought of as an application of ``prompt engineering'' in which the researchers attempts to identify patterns in LLM misclassifications and change the codebook to correct any consistent misperceptions. For each task, we did at most one round of codebook updates. To measure the effect that codebook updates had on LLM labeling, we re-label the training data subsets using the final codebooks. 

\begin{figure}[H]
    \centering
    \includegraphics[width = .9 \textwidth]{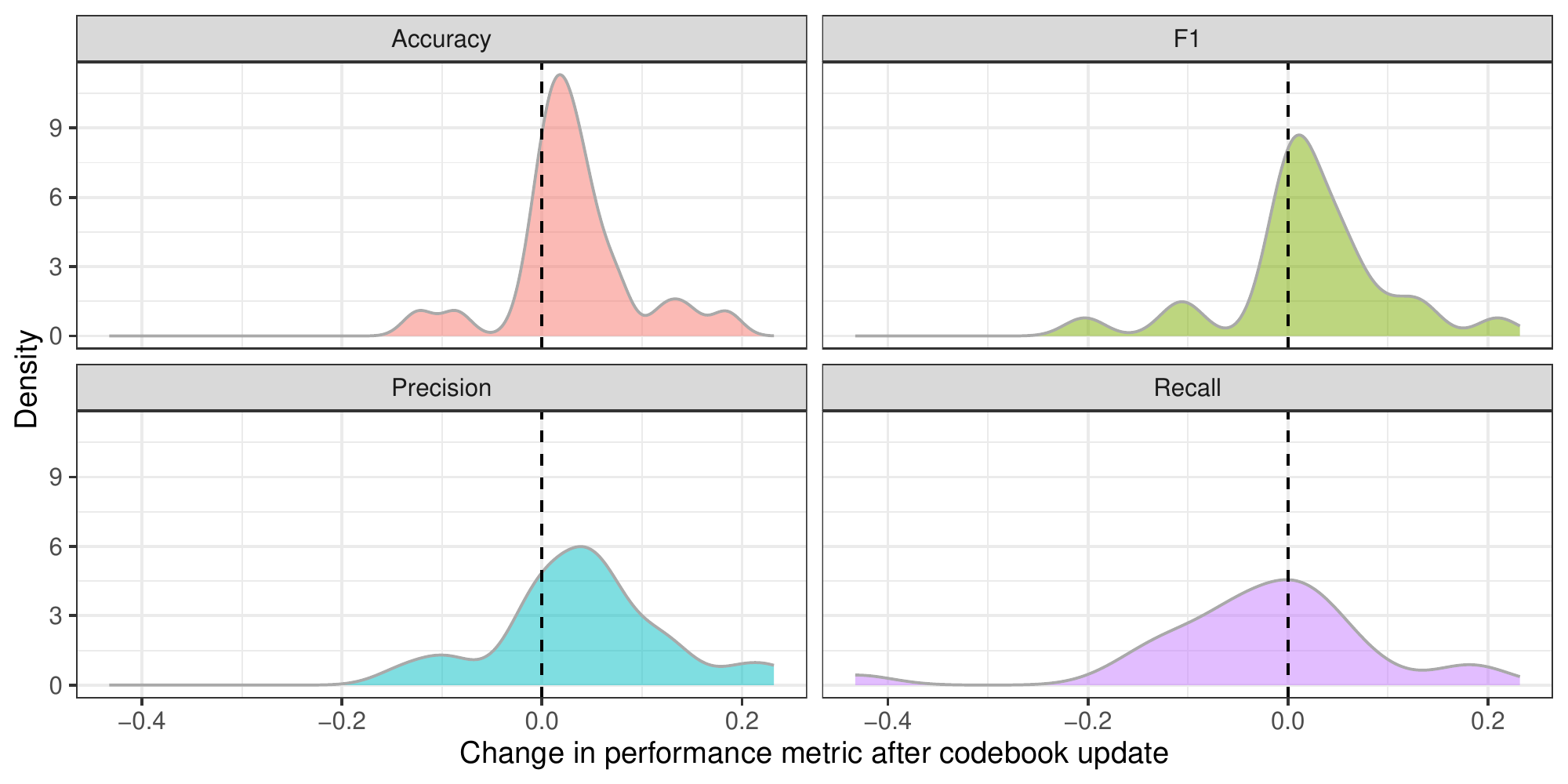}
    \caption{Change in LLM annotation performance on training data after one round of codebook updates.}
    \label{fig:codebook_updates}
\end{figure}

Figure \ref{fig:codebook_updates} shows the distributions of change in performance metrics after updating the codebook. This analysis demonstrates whether and how the codebook update process affects LLM annotation, holding constant the data and conceptual categories. In most cases, the codebook update process led to modest improvement in accuracy and F1. Recall decreased in more cases than improved after codebook updates. Precision, on the other hand, improved in a majority of cases, driving the improvement in accuracy and F1. Given these results, the subtleties of prompt construction do not appear to be a significant lever on performance. Still, although the magnitude of improvement was generally small, researchers experiencing subpar LLM classification performance on their text data can use human-in-the-loop codebook refinement to ensure that their instructions are not to blame. 

\section{Use cases}

If an LLM achieves satisfactory performance on domain-specific benchmarks for a given annotation task, there are numerous ways that the LLM could be integrated into a text analysis project. How researchers augment their project with an LLM may depend on the LLM performance against the human labeled data, their budget, the size of the dataset, and the availability of human annotators. Table \ref{tab:cases} displays four potential use cases.  

\begin{table} 
\singlespacing
    \centering
    \begin{tabular}{p{5cm}|p{11cm}}
         \toprule
         \textbf{Use case} & \textbf{Description} \\
         \midrule
         1) Confirming the quality of human-labeled data & If a researcher has data that is already labeled by human annotators, an LLM may augment the text analysis process by confirming the quality of the human labels. If LLM performance against the human labels is high, this signals that both the LLM and humans made similar annotation decisions. If LLM performance is low, this indicates that either the humans, the LLM, or both made mistakes during the annotation process.  \\
         \midrule
         2) Identifying cases to prioritize for human review & A researcher can manually review annotations with consistency scores lower than 1.0. Additionally, if the particular LLM task achieved high recall, then the researcher may use the LLM to identify potential positive cases in previously unseen data. Then, human annotators can manually review all positive-labeled cases. \\
         \midrule
         3) Producing labeled data to finetune and validate a supervised classifier & There are various situations in which a researcher may use an LLM to procure training data to finetune a supervised classifier for labeling their corpus.  \\
         \midrule
         4) Classifying the entire corpus directly & In the simplest case, a researcher may find LLM performance to be satisfactory and opt to use the LLM to classify the entire remaining corpus.  \\
         \bottomrule
    \end{tabular}
    \caption{Use cases for automated annotation with LLMs.}
    \label{tab:cases}
\end{table}

If the LLM's performance on some or all dimensions is poor, then the researcher can continue to refine the codebook using new labeled data or abandon the use of LLMs for the dimensions with unsatisfactory performance. If researchers opt to continue using an LLM despite initial poor performance, one strategy to simplify complex annotations may be disaggregation. For instance, if a researcher finds poor annotation performance on the dimension of ``hate speech,'' they may achieve better performance by disaggregating ``hate speech'' into component indicators (such as threats, slurs, stereotyping, and so on) and adding each of the separate indicators as new dimensions into the codebook, then beginning again at Step 1 in the workflow.

\section{Conclusion}

We observe significant heterogeneity in LLM performance across a range of social science annotation tasks. Performance depends on both attributes of the text and the conceptual categories being measured.  For instance, whether a classification task is a factual question about a text sample or requires judgement has significant bearing on classification strategy \citep{Balagopalan23}. To address the myriad sources of variation in annotation performance, we recommend a flexible approach to LLM-augmented annotation that foregrounds human annotation. 

To address these challenges, we present an original workflow for augmenting text annotation with an LLM along with several use cases for this workflow. We validate the workflow by replicating 27 annotation tasks taken from 11 social science articles published in high-impact journals. We find that LLMs can offer high-quality labels on a wide variety of tasks for a fraction of the cost and time of alternative options, such as crowd-sourced workers and undergraduate research assistants. Still, it is imperative that researchers validate the performance of LLMs on a task-by-task basis, as we find significant heterogeneity in performance, even across tasks within a single dataset.

\clearpage

\singlespacing
\bibliographystyle{apalike}
\bibliography{bib.bib}

\clearpage

\clearpage

\section*{Appendix}

\setcounter{table}{0}
\renewcommand{\thetable}{A\arabic{table}}

\singlespacing
\begin{table}[H] 
    \centering
    \begin{tabular}{p{3cm}p{7cm}p{4cm}p{1cm}}
         \toprule
         Author(s) & Title & Journal & Year \\
         \midrule
         Gohdes & Repression Technology: Internet Accessibility and State Violence & American Journal of Political Science & 2020 \\
         \midrule
         Hopkins, Lelkes, and Wolken & The Rise of and Demand for Identity-Oriented Media Coverage & American Journal of Political Science & (R\&R) \\
         \midrule
         Schub & Informing the Leader: Bureaucracies and International Crises & American Political Science Review & 2022 \\
         \midrule
         Busby, and Gubler, Hawkins & Framing and blame attribution in populist rhetoric & Journal of Politics & 2019 \\
         \midrule
         Müller & The Temporal Focus of Campaign Communication & Journal of Politics & 2021 \\
         \midrule
         Cusimano and Goodwin &  People judge others to have more voluntary control over beliefs than they themselves do & Journal of Personality and Social Psychology & 2020 \\
         \midrule
         Yu and Zhang & The Impact of Social Identity Conflict on Planning Horizons & Journal of Personality and Social Psychology & 2022 \\
         \midrule
         Card et al. & Computational analysis of 140 years of US political speeches reveals more positive but increasingly polarized framing of immigration & PNAS & 2022 \\
         \midrule
         Peng, Romero, and Horvat & Dynamics of cross-platform attention to retracted papers & PNAS & 2022 \\
         \midrule
         Saha  et al. & On the rise of fear speech in online social media \tablefootnote{This article uses the Gab hate speech corpus \citep{kennedy22}. We include \cite{Saha23} here, rather than the original source of the labeled data, to emphasize the application of these data in applied social science research.} & PNAS & 2022 \\
         \midrule
         Wojcieszak et al. & Most users do not follow political elites on Twitter; those who do show overwhelming preferences for ideological congruity & Science Advances & 2022 \\
         \bottomrule
         & 
    \end{tabular}
    \caption{Sources of annotation tasks that replicated in analysis.}
    \label{tab:articles}
\end{table}

\begin{table}[]
\centering
\begin{tabular}{p{3cm}p{12cm}}
\toprule
\multicolumn{1}{c}{Study}         & \multicolumn{1}{c}{Annotation tasks}                                                                                                                                                                     \\ \midrule
Gohdes (2020)                     & Code Syrian death records for specific type of killing: targeted or untargeted                                                                                                                          \\ \midrule
Hopkins, Lelkes, \& Wolken (2023) & Coding headlines, tweets, and Facebook share blurbs to identify references to social groups defined by a) race/ethnicity; b) gender/sexuality; c) politics; d) religion \\ \midrule
Schub (2020)                      & Code presidential-level deliberation texts from the Cold War as political or military                                                                                                                   \\ \midrule
Busby, Gubler, \& Hawkins (2019)  & Code open-ended responses for three rhetorical elements: attribution of blame to a specific actor, the attribution of blame to a nefarious elite actor, and a positive mention of the collective people \\ \midrule
Müller (2021)                     & Code sentences from party manifestos for temporal direction: past, present, or future                                                                                                                   \\ \midrule
Cusimano \& Goodwin (2020)        & Code respondents' written statements on climate change for the presence of either (a) generic reasoning about beliefs or (b) supporting evidence for the belief                                         \\ \midrule
Yu \& Zhang (2023)                & Code respondents' plans for the future into two categories:proximate future and distant future                                                                                                          \\ \midrule
Card et al. (2022)                & Code congressional speeches for whether they are about immigration, along with an accompanying tone: proimmigration, antiimmigration, or neutral                                                        \\ \midrule
Peng, Romero, \& Horvat (2022)    & Code whether tweets express criticism with respect to the findings of academic papers                                                                                                                   \\ \midrule
Saha et al. (2020)                & Code Gab posts as a) fear speech, b) hate speech, or c) normal. Further, a post could have both fear and hate components, and,thus, these were annotated with multiple labels                           \\ \midrule
Wojcieszak et al. (2020)          & Code whether a quote tweet was negative, neutral or positive toward the message and/or the political actor, independently of the tone of the original message                                           \\ \bottomrule
\end{tabular}
\caption{Descriptions of annotation tasks replicated in analysis.}
\label{tab:articles2}
\end{table}

\end{document}